\documentclass[letterpaper, 10 pt, conference]{ieeeconf}

\IEEEoverridecommandlockouts
\usepackage{makecell}
\usepackage{svg}
\usepackage{algorithm}
\usepackage{algpseudocode}
\makeatletter
\def\caption@documentclass{standard}
\makeatother
\usepackage{subcaption}
\usepackage{cite}
\usepackage{graphicx}
\usepackage{amsmath,amssymb}
\usepackage{adjustbox}
\usepackage{booktabs}

\begin{document}

\title{\LARGE \bf
Learning Event-based Obstacle Avoidance via Differentiable Simulation
}

\author{
Chenlu Shou \qquad Yunfan Ren\qquad Jiaxu Xing \qquad Davide Scaramuzza
\thanks{This work was supported by the Government of Singapore’s Technology Development Program through the Singapore University of Technology and Design (SUTD), European Union's Horizon Europe Research and Innovation Program under grant agreement No.~101120732 (AUTOASSESS) and the European Research Council (ERC) under grant agreement No.~864042 (AGILEFLIGHT). }
\thanks{Corresponding author: Yunfan Ren. The authors are with the Robotics and Perception Group, University of Zurich, Switzerland ({\tt\small https://rpg.ifi.uzh.ch/}).}
}

\maketitle
\thispagestyle{empty}
\pagestyle{empty}

\begin{abstract}
Reliable obstacle avoidance is essential for safe quadrotor flight in unknown environments, yet conventional frame-based cameras remain challenged by motion blur, limited dynamic range, and latency under fast motion.
Event cameras, with sub-millisecond latency, high dynamic range, and negligible motion blur, offer a promising alternative, but leveraging them for closed-loop control remains difficult.
Existing approaches either adopt cascaded architectures that first estimate depth and then map it to control commands, introducing error propagation, or achieve end-to-end control through costly offline pretraining before online fine-tuning can even begin.
We address these challenges with a CUDA-based event simulator that exploits GPU parallelism to render events efficiently within the training loop, integrated into a differentiable simulation pipeline that enables direct gradient backpropagation through the differentiable physics model, improving sample efficiency.
This allows us to train an end-to-end policy that maps event frames directly to control commands, without intermediate depth estimation.
We validate our approach through zero-shot real-world deployment on a quadrotor, achieving robust obstacle avoidance in both odometry-assisted and odometry-free settings, including under low-light conditions where conventional cameras typically fail.
\end{abstract}

\section{Introduction}
\label{sec:intro}
Quadrotors have been widely adopted in autonomous robotics for applications such as industrial inspection, package delivery, and search-and-rescue, where reliable obstacle avoidance is essential for safe operation in unknown environments~\cite{xing2023autonomous}.
However, achieving robust vision-based obstacle avoidance remains challenging, particularly during high-speed flight, because conventional frame-based cameras are susceptible to motion blur, limited dynamic range, and sensing latency.
Unlike traditional cameras that capture image frames at a fixed rate, event cameras \cite{eventvision}, modeled after the biological retina, operate asynchronously.
Each pixel functions independently, and an event is triggered only when the change in log-intensity at pixel coordinates exceeds a predefined threshold, with polarity indicating an increase or decrease in brightness.
This bio-inspired sensing mechanism inherently minimizes motion blur and provides sub-millisecond latency and a much higher dynamic range, making event cameras highly suitable for agile applications \cite{dur2026outplaying, falanga2020dynamic, 10161392}.
Most notably, Sony AI's Ace \cite{dur2026outplaying}, the first real-world table tennis AI agent competitive with human athletes, uses event-based spin estimation to return high-speed shots.

In this work, we leverage these advantages to address vision-based obstacle avoidance in quadrotor flight.
A closely related method \cite{bhattacharya2025monocular} adopts a cascaded architecture, in which a perception network first predicts depth from the input event stream, and a separate velocity predictor maps this estimated depth to control commands.
This architecture makes control performance dependent on the accuracy of an intermediate depth estimate: errors in depth prediction can propagate to the control commands and compromise obstacle avoidance.
This dependence motivates learning a direct mapping from event data to control commands, allowing the visual representation to be optimized jointly with the control policy for the obstacle-avoidance task.
\begin{figure}
    \centering
\includegraphics[width=\linewidth]{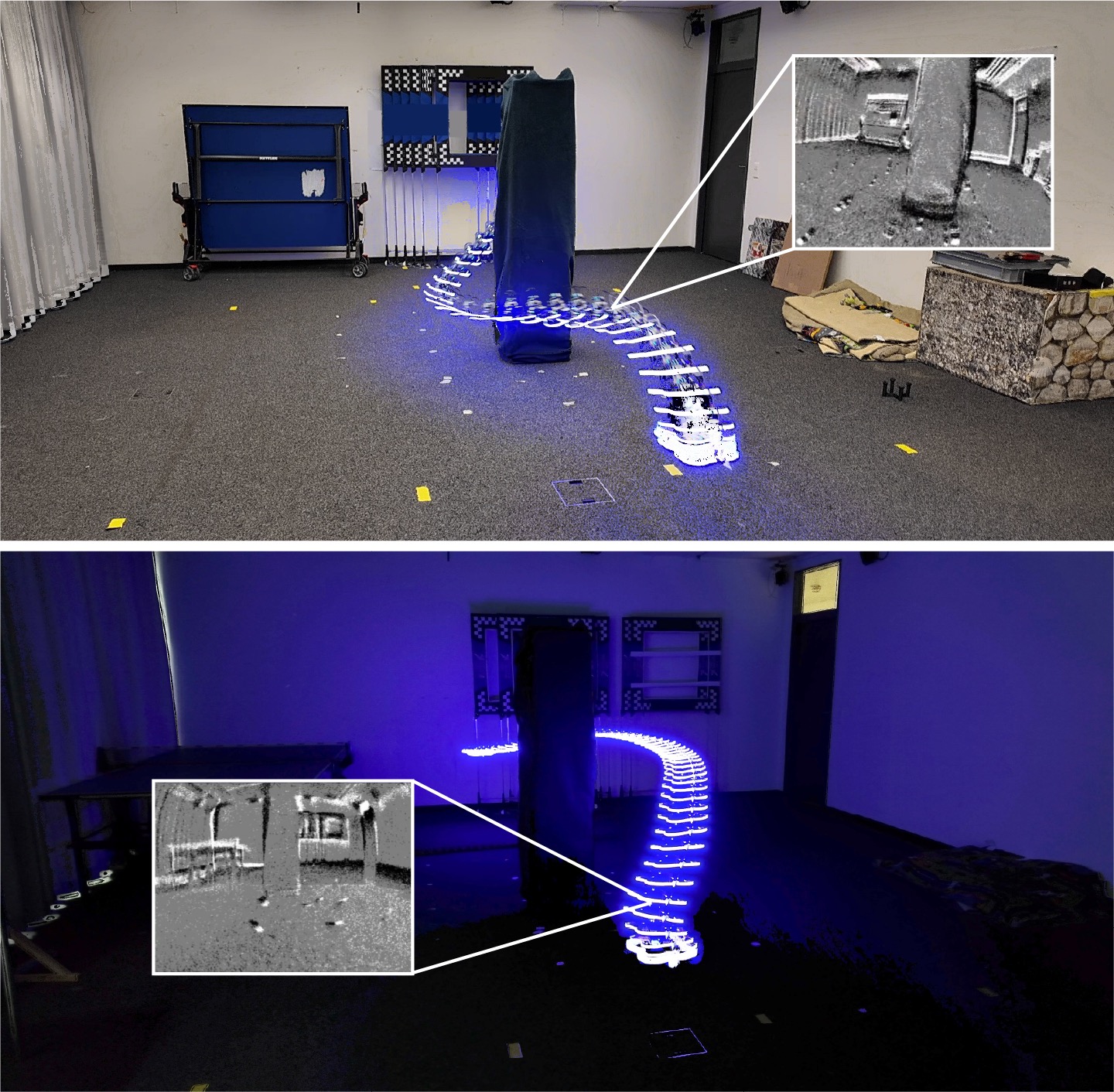}
    \caption{Real-world flight trajectories demonstrating event-based obstacle avoidance in an odometry-free setting (top) and under challenging \emph{low-light} conditions (bottom). The inset in each panel shows the onboard event-camera view at a representative time step.}
    \vspace{-0.5cm}
    \label{fig:image1}
\end{figure}
An end-to-end policy that maps event data directly to control commands would avoid this issue, but training such a policy at scale requires an event simulator that can generate large volumes of event data efficiently.
Existing event generation tools remain computationally expensive, making them a major bottleneck for policy training.
\begin{figure*}
    \centering
    \includegraphics[width=0.98\textwidth]{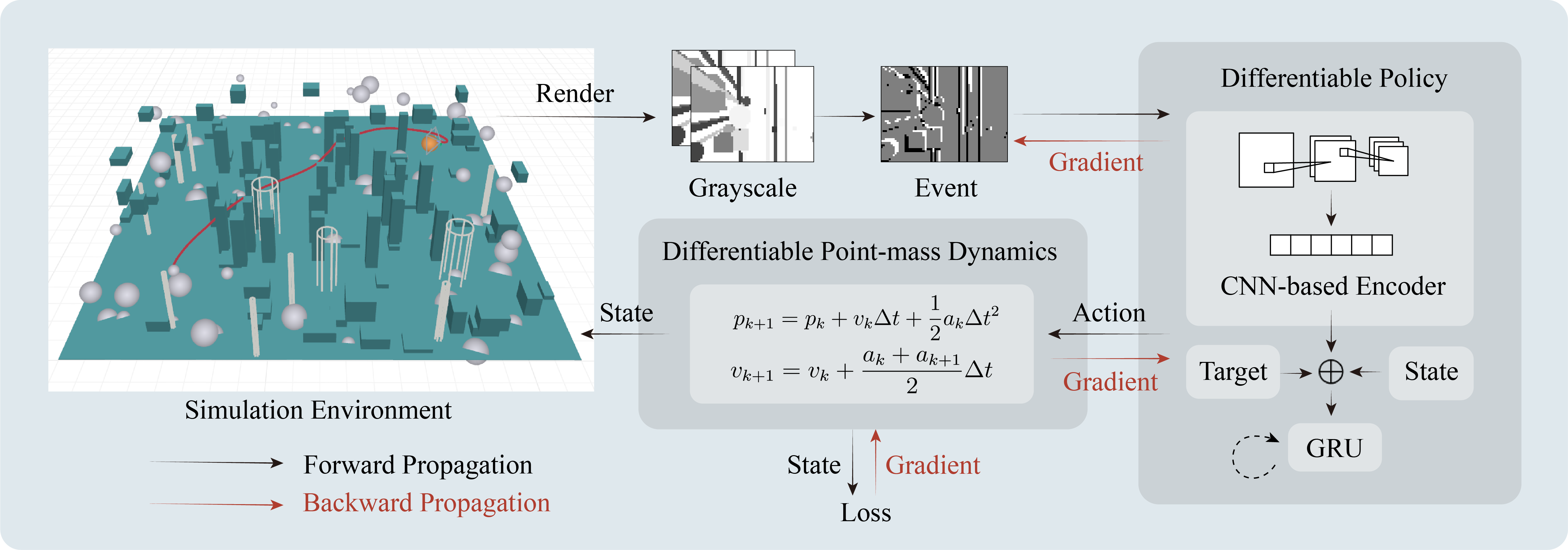}
    \caption{System Framework Overview. The simulator renders a grayscale frame from each quadrotor state, and two consecutive frames are used to compute the event frame. The event frame is fed into a differentiable policy consisting of a CNN and a GRU, which predicts an action. The differentiable point-mass dynamics model then propagates the current state forward using this action. A loss is computed based on the resulting state and action, and the corresponding gradients are propagated backward through the entire differentiable pipeline to update the policy.}
    \vspace{-0.2cm}
    \label{fig:framework}
\end{figure*}

Furthermore, using model-free reinforcement learning (RL), such as proximal policy optimization (PPO) \cite{ppo2017}, to train such a policy is notoriously sample-inefficient, particularly when learning directly from high-dimensional visual inputs such as event data, which require large numbers of interactions to extract useful features.
Differentiable simulation offers a promising alternative.
By computing gradients through the simulated dynamics analytically, policies can be optimized via backpropagation, improving sample efficiency. Recent works have demonstrated the effectiveness of differentiable simulation across various learning scenarios \cite{zhang2025learning, pan2026learning, opticalflow}.
In this work, we present an end-to-end control framework for event-based obstacle avoidance that maps event frames directly to control commands.
We investigate both odometry-assisted and odometry-free settings to examine the policy's dependence on external velocity estimates.
In the latter, the policy uses only onboard attitude information as state input and learns to infer velocity-related information from the event history.

To summarize, the key contributions of this work are:
\begin{itemize}
\item  We develop a CUDA-based event simulator that exploits GPU parallelism to generate event observations efficiently within the policy training loop.
At a resolution of $64 \times 48$ with 128 parallel environments, our simulator achieves 60,248 environment steps per second, approximately $14\times$ the throughput of the state-of-the-art~\cite{rebecq2018esim}.
We further assess the transfer of the simulated event representation through offline velocity estimation on real-world event data.

\item  We integrate this simulator into a differentiable simulation pipeline that optimizes the control policy by backpropagating gradients through the simulated dynamics.
This framework enables policy convergence in approximately 3 hours on a laptop equipped with an NVIDIA RTX 5060 GPU, without a separate offline event-data generation and representation-pretraining stage.

\item We demonstrate zero-shot deployment on a real-world quadrotor, achieving obstacle avoidance in both odometry-assisted and odometry-free settings.
We further demonstrate successful flight under low-light conditions with illuminance as low as 5.8 lux.

\end{itemize}

\section{Related Work}

\subsection{Event Simulation}
Simulating event cameras is essential for training learning-based policies that use events as input, since real event data is hard to collect at scale.
Early work \cite{mueggler2017event} simulates events by rendering images at a fixed, densely-sampled framerate along the camera trajectory and linearly interpolating pixel-wise brightness between samples to estimate event timestamps.
ESIM \cite{rebecq2018esim} improves on this by tightly coupling the rendering engine with the event simulator, adaptively sampling frames based on the predicted rate of brightness change or pixel displacement, so that new frames are rendered only when necessary.
However, its rendering frequency is still limited when using the OpenGL rendering engine on a laptop, and even slower with the photorealistic rendering engine, making it computationally expensive to use inside a large-scale policy training loop. Other approaches, such as Vid2E \cite{gehrig2020video}, convert existing video datasets into event streams, but rely on pre-recorded or rendered videos rather than real-time simulation.
Therefore, it is necessary to design a customized, efficient event data simulator suitable for policy training.

\subsection{Obstacle Avoidance via Conventional Perception}
Obstacle avoidance for aerial robots has predominantly relied on
conventional frame-based cameras.
Classical pipelines map the environment from depth measurements and plan trajectories over this representation \cite{tordesillas2021faster}.
To avoid the associated mapping latency, some methods, e.g.,~\cite{zhou2020ego} reduce reliance on the explicit map and instead derive obstacle gradients directly from local obstacle information, while fully mapless approaches act directly on instantaneous depth via a library of motion primitives \cite{florence2020integrated}.
In contrast, other work bypasses explicit geometry, learning policies that map visual input to control through imitation learning from human or expert demonstrations \cite{loquercio2018dronet,xing2024bootstrapping,ross2013learning},
self-supervised interaction with real-world collisions
\cite{gandhi2017learning}, or differentiable simulation of a depth-to-control policy \cite{zhang2025learning}.
An alternative approach trains an end-to-end policy on optical flow simulated from depth maps~\cite{opticalflow}.
These camera-based methods nonetheless remain constrained by the imaging process itself, continuing to suffer from motion blur, limited dynamic range, and high latency under the challenging condition of agile quadrotor flight.
Beyond cameras, active ranging sensors offer complementary trade-offs.
For example, LiDAR provides accurate long-range depth at added weight and power \cite{ren2025safety}, whereas radar provides coarser spatial resolution \cite{wessendorp2021obstacle}.
\subsection{Event-based Drone Control}
Event cameras offer a fundamentally different sensing paradigm that directly addresses the limitations of conventional cameras.
Several works have explored event cameras for vision-based tasks in quadrotors.
Early efforts to combine event cameras with learning-based control are validated entirely in simulation, without addressing the challenges of real-world deployment \cite{vemprala2021representation, salvatore2020neuro}.
For real-world applications, Falanga et al. \cite{falanga2020dynamic} demonstrate low-latency dynamic obstacle dodging on an autonomous quadrotor using onboard event vision.
Andersen et al. \cite{andersen2022event} apply an event camera in racing gate detection using a sparse gated recurrent network, and validate the approach in a real-world drone racing task, while Vitale et al. \cite{vitale2021event} present an SNN-based line-tracking controller on a constrained dualcopter platform. For static-obstacle avoidance tasks during drone flight, the first method presented by Bhattacharya et al. \cite{bhattacharya2025monocular} uses a monocular event camera with a cascaded depth and velocity prediction pipeline.
As discussed in Sec.~\ref{sec:intro}, errors in the intermediate depth prediction can propagate to the final control output.
Messikommer et al.
\cite{messikommer2026approximate} instead train an end-to-end event-based policy through an approximate imitation learning framework.
This method combines offline training to learn a task-specific event representation space with online training to fine-tune the action decoder, thus avoiding costly event rendering.
However, this approach still depends on a computationally expensive offline stage to render events and pretrain the policy, taking on the order of 200 hours before online fine-tuning can begin.
In contrast, our method proposes an efficient event simulator that enables end-to-end policy optimization without a separate offline rendering and pretraining stage.

\section{Methodology}
The overview of our system framework is summarized in Fig.~\ref{fig:framework}, illustrating the differentiable pipeline from event rendering to policy inference and dynamics propagation, along with the flow of gradients back-propagated through the differentiable pipeline.
We will introduce our differentiable learning framework along with event simulator in detail next.
\subsection{Problem Formulation}
We model the quadrotor's motion as a discrete-time dynamical system.
At time step $t$, let $x_t \in \mathcal{X}$ denote its state and $u_t \in \mathcal{U}$ denote the control input, where $\mathcal{X}$ and $\mathcal{U}$ are the state and control spaces, respectively.
The current state and control input determine the next state through
$
    x_{t+1} = f(x_t, u_t),
$
where $f$ describes the quadrotor dynamics.
Using this model, we formulate obstacle avoidance as a finite-horizon control problem.

The quadrotor receives an observation $o_t \in \mathcal{O}$ at each time step, obtained via the observation function $h: \mathcal{X} \rightarrow \mathcal{O}$, such that $o_t = h(x_t)$.
The control policy $\pi_\theta$, parameterized as a neural network, maps the observation to a control input, $u_t = \pi_\theta(o_t)$.
At each time step, a cost $l_t = l(x_t, u_t)$ is used to measure the stage loss based on the current state $x_t$ and the predicted control input $u_t$.
The policy optimization problem then amounts to identifying the parameters $\theta^*$ that minimize the accumulated cost over a rollout of horizon $N$:
\begin{align}
\min_{\theta} \mathcal{L}_\theta &= \sum_{t=0}^{N-1} l(x_t, u_t) = \sum_{t=0}^{N-1} l(x_t, \pi_\theta(o_t)) \\
\theta &\leftarrow \theta - \eta \nabla_\theta \mathcal{L}_\theta
\end{align}
where $\eta$ denotes the learning rate.

\subsection{Event Rendering}
Training an end-to-end policy requires large amounts of event data, which are costly to collect through real-world flight.
Simulation provides a scalable alternative, but event generation can become a computational bottleneck within the policy training loop.
To address this bottleneck, we develop a CUDA-based event simulator that exploits GPU parallelism to generate event observations efficiently.

An event is represented as $e_k = (t_k, x_k, y_k, p_k)$, where $t_k$ is its timestamp, $(x_k, y_k)$ is its pixel location, and $p_k \in \{+1, -1\}$ indicates an increase or decrease in log-intensity.
During flight, camera motion causes obstacle boundaries to move across the image, producing brightness changes that trigger events.
These boundaries provide important visual cues about the location and extent of obstacles, helping the policy identify potential collisions.
Our simulator therefore focuses on these boundary-induced changes and directly generates event frames at the policy's control rate.
Specifically, we assign each obstacle a uniform grayscale intensity and detect intensity changes between frames rendered at consecutive camera poses.
This approximation allows us to generate event frames efficiently from obstacle geometry.

\subsubsection{Ray Casting}
We approximate obstacles using simple geometric primitives and assign each obstacle a uniform, randomly sampled grayscale intensity.
This simplifies surface appearance while retaining obstacle boundaries and occlusions that provide useful cues for obstacle avoidance.
Given the camera pose, we cast a ray through each pixel and compute its intersections with the obstacle geometry, including spheres, voxels, and vertical and horizontal cylinders.
The pixel is assigned the grayscale intensity of the nearest intersected obstacle, accounting for visibility and occlusion.
These geometric primitives admit closed-form ray-intersection solutions, allowing each pixel to be rendered independently and efficiently in parallel on the GPU.

\subsubsection{Event Generation}
To balance physical fidelity with rendering efficiency, events are synthesized directly by comparing two consecutive grayscale frames, $\text{Grayscale}_t$ and $\text{Grayscale}_{t-1}$, rendered at successive control time steps, as an efficient approximation of the underlying per-pixel intensity change. For each pixel, the polarity $p_k$ is computed as
\begin{equation}
    p_k =
    \begin{cases}
        +1, & \text{if } \text{Grayscale}_t - \text{Grayscale}_{t-1} > C, \\
        -1, & \text{if } \text{Grayscale}_t - \text{Grayscale}_{t-1} < -C, \\
        0, & \text{otherwise},
    \end{cases}
\end{equation}
where $C$ denotes the contrast threshold. This formulation provides a fast approximate representation of event generation.

\subsubsection{Low-Resolution Rendering}
\begin{figure}[t]
    \centering
    \begin{subfigure}[b]{1.0\linewidth}
        \centering
        \includegraphics[width=1.0\linewidth]{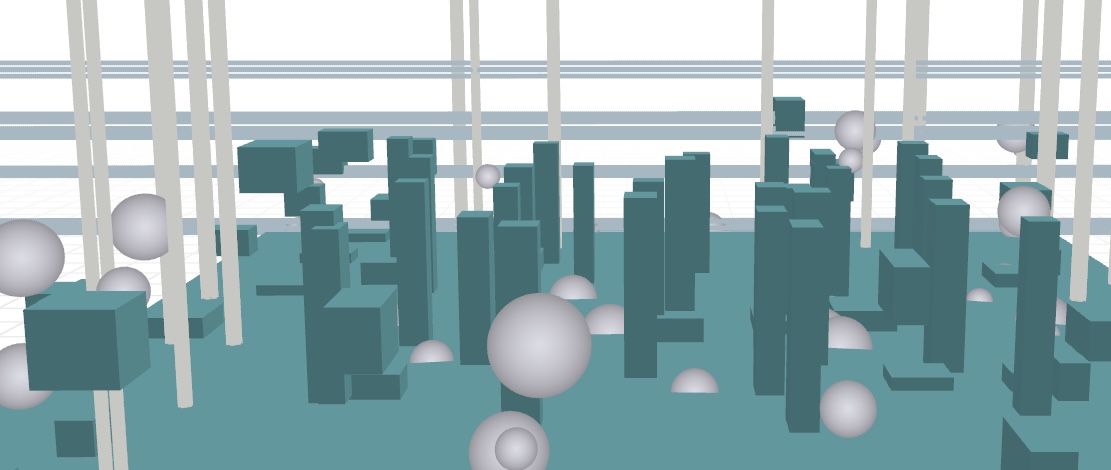}
        \caption{Simulation Environment}
        \label{fig:placeholder}
    \end{subfigure}

    \vspace{0.5em}

    \begin{subfigure}[b]{0.32\linewidth}
        \centering
        \includegraphics[width=\linewidth]{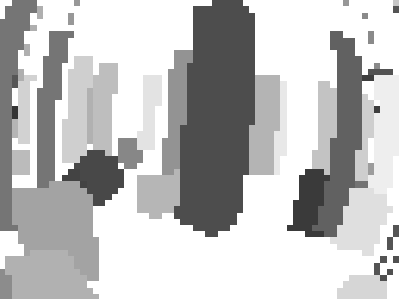}
        \caption{Grayscale $t-1$}
        \label{fig:grayscale2}
    \end{subfigure}
    \hfill
    \begin{subfigure}[b]{0.32\linewidth}
        \centering
        \includegraphics[width=\linewidth]{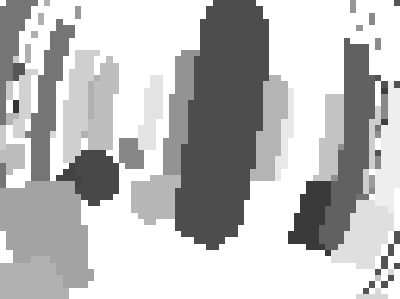}
        \caption{Grayscale $t$}
        \label{fig:grayscale1}
    \end{subfigure}
    \hfill
    \begin{subfigure}[b]{0.32\linewidth}
        \centering
        \includegraphics[width=\linewidth]{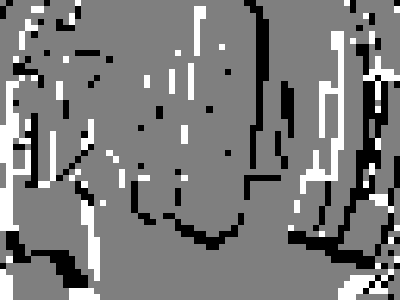}
        \caption{Event}
        \label{fig:event $t$}
    \end{subfigure}
    \caption{Illustration of the simulation environment along with example rendered outputs. (a) shows the simulation environment, (b)-(c) the rendered grayscale frame, and (d) the resulting event frame synthesized by comparing consecutive grayscale frames.}
    \label{fig:pipeline}
    \vspace{-0.2cm}
\end{figure}

We render event frames at a low spatial resolution ($64\times 48$), which offers two key benefits.
First, it enables more efficient training by reducing the computational cost of simulation.

Second, during deployment, real event camera frames (e.g., $640\times 480$) are down-sampled to low resolution to align with our simulator before being fed to the policy.
This process acts as an implicit low-pass filter that suppresses sensor noise and task-irrelevant details in real event frames. These properties allow our event simulator to achieve high rendering efficiency while narrowing the sim-to-real gap.

\subsubsection{Lens Distortion}
First-person view (FPV) quadrotors are usually equipped with wide field-of-view (FOV) cameras to capture a broader view of the surroundings during high-speed flight, but they often introduce noticeable lens distortion. Therefore, we simulate this effect within our rendering pipeline using the Kannala-Brandt fisheye model to further narrow the sim-to-real gap. Given the undistorted incidence angle $\phi$ of a camera ray, the corresponding distorted angle $\phi_d$ is computed as,
\begin{equation}
    \phi_d = \phi\left(1 + k_1\phi^2 + k_2\phi^4 + k_3\phi^6 + k_4\phi^8\right),
\end{equation}
where $k_1, k_2, k_3, k_4$ are distortion coefficients obtained by calibrating the real camera.

\subsubsection{Noise Simulation}
Since real event cameras are inherently sensitive to noise, we explicitly model it as fluctuations in the contrast threshold and spurious background activity. Following~\cite{rebecca2019high}, rather than using a fixed contrast threshold, we independently resample the positive and negative thresholds at every simulation timestep from clipped Gaussian distributions,

\begin{equation}
    C = \text{clip}\big(\tilde{C},\, C_{\min},\, C_{\max}\big), \quad \text{where } \tilde{C} \sim \mathcal{N}(\mu, \sigma_C^2)
\end{equation}

In addition, we inject background activity noise by randomly triggering events at a small fraction of pixels, independent of scene edges, emulating the noise commonly observed in dark or low-contrast regions of real event cameras.

Fig.~\ref{fig:pipeline} illustrates this rendering pipeline, from the simulated environment to the synthesized event output.

\subsection{Learning Framework}
\subsubsection{Policy Optimization via Differentiable Simulation}

To optimize the control policy, we leverage differentiable simulation for gradient backpropagation.
This framework allows analytical calculation of gradients through a differentiable physics model, significantly improving sample efficiency.

The policy gradient is computed via Back-propagation Through Time (BPTT) $\nabla_{\theta} \mathcal{L}_\theta=$
\begin{equation}
\frac{1}{N} \sum_{t=0}^{N-1} ( \sum_{i=0}^{t} \frac{\partial l_t}{\partial x_t} \prod_{j=i+1}^{t} \left( \frac{\partial x_j}{\partial x_{j-1}} e^{-\alpha \Delta t} \right) \frac{\partial x_i}{\partial \theta} + \frac{\partial l_t}{\partial u_t} \frac{\partial u_t}{\partial \theta} )
\end{equation}
where $\alpha$ denotes the temporal gradient decay factor.

Rather than employing an exact quadrotor dynamics model, Zhang et al. \cite{zhang2025learning} demonstrate that a simplified point-mass approximation can effectively transfer to real-world platforms.
Inspired by their work, we adopt a point-mass differentiable simulator to model the quadrotor dynamics, where position and velocity are updated via numerical integration:
\begin{equation}
\begin{aligned}
    \mathbf{v}_{t+1} &= \mathbf{v}_t + \frac{\mathbf{a}_t + \mathbf{a}_{t+1}}{2} \Delta t \\
    \mathbf{p}_{t+1} &= \mathbf{p}_t + \mathbf{v}_t \Delta t + \frac{1}{2} \mathbf{a}_t \Delta t^2
\end{aligned}
\end{equation}
\begin{figure*}[t]
  \centering

  \begin{subfigure}[b]{0.49\textwidth}
    \centering
    \begin{minipage}[b]{0.49\linewidth}
      \centering
      \includegraphics[width=\linewidth]{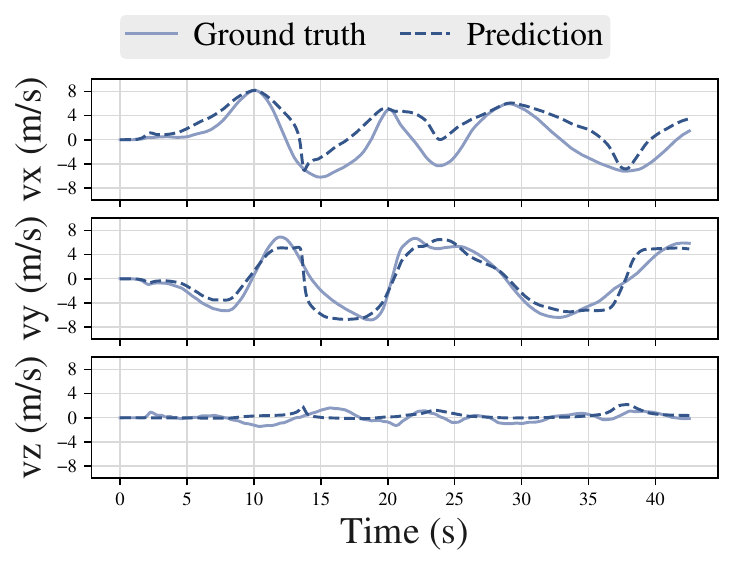}\\[1pt]
      \small Velocity prediction
    \end{minipage}%
    \hfill
    \begin{minipage}[b]{0.49\linewidth}
      \centering
      \includegraphics[width=\linewidth]{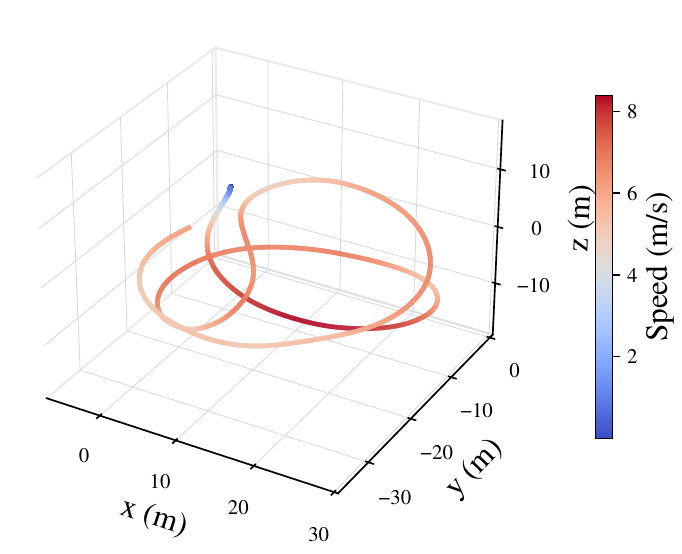}\\[1pt]
      \small Flight trajectory
    \end{minipage}
    \caption{Sequence 1}
    \label{fig:sub-a}
  \end{subfigure}%
  \hfill
  \begin{subfigure}[b]{0.49\textwidth}
    \centering
    \begin{minipage}[b]{0.49\linewidth}
      \centering
      \includegraphics[width=\linewidth]{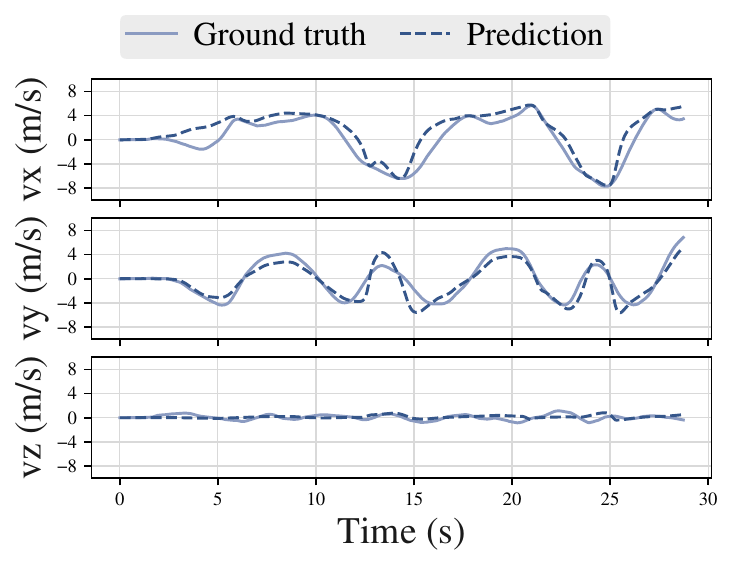}\\[1pt]
      \small Velocity prediction
    \end{minipage}%
    \hfill
    \begin{minipage}[b]{0.49\linewidth}
      \centering
      \includegraphics[width=\linewidth]{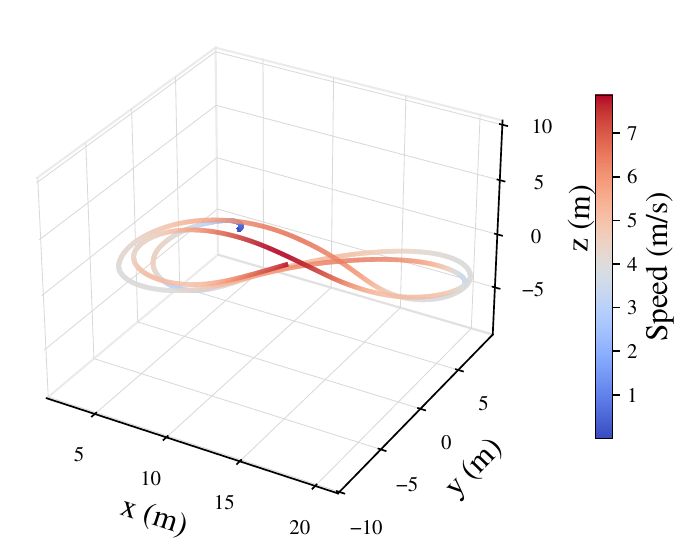}\\[1pt]
      \small Flight trajectory
    \end{minipage}
    \caption{Sequence 2}
    \label{fig:sub-b}
  \end{subfigure}

  \caption{Offline evaluation of the velocity estimator on two representative real-world sequences from the UZH-FPV Drone Racing Dataset~\cite{Delmerico19icra}. For each sequence, the left plots show the predicted versus ground-truth velocity over time, and the right plot shows the corresponding flight trajectory.}
  \label{fig:offline_eval}
  \vspace{-0.5cm}
\end{figure*}
\subsubsection{Network Architecture}
Our policy $\pi_\theta$ is parameterized as a neural network comprising a convolutional neural network (CNN)-based encoder for event feature extraction and a gated recurrent unit (GRU) layer for temporal memory.
The extracted visual features are fused with the quadrotor's state as well as the target velocity, directed toward the goal, to predict acceleration commands.
The state vector incorporates attitude and velocity in odometry-assisted settings, but only attitude in odometry-free ones.
The entire network is designed to be lightweight for real-time control inference on the quadrotor's onboard computer.

\subsubsection{Loss function}
The total loss function consists of velocity tracking loss, velocity estimation loss, obstacle avoidance loss, and control smoothness loss, each weighted by a corresponding coefficient $\lambda$ (omitted in Eq.~\ref{eq:loss_total} for clarity):
\begin{equation}
\label{eq:loss_total}
    \mathcal{L} = \mathcal{L}_{\text{vel}} + \mathcal{L}_{\text{est}} + \mathcal{L}_{\text{obs}} + \mathcal{L}_{\text{acc}} + \mathcal{L}_{\text{jerk}}
\end{equation}

We define $\mathcal{L}_\text{vel}$ using the Smooth $L_1$ (Huber) loss to encourage the quadrotor's moving average velocity to track the commanded target velocity:
\begin{equation}
    \mathcal{L}_\text{vel} = \frac{1}{T} \sum_{t=1}^{T} \text{SmoothL1}\left( \left\| \bar{\mathbf{v}}_t - \mathbf{v}_{t}^{\text{des}} \right\|_2, 0\right)
\end{equation}
where $\bar{\mathbf{v}}_t$ denotes the moving average velocity computed over a temporal window, introduced to mitigate the sensitivity to velocity jitter, and $\mathbf{v}_{t}^{\text{des}}$ is the desired target velocity.

To enable navigation in odometry-free scenarios, $\mathcal{L}_{\text{est}}$ supervises the network's self-estimated velocity against the ground-truth velocity obtained from the simulator, penalized using a mean squared error (MSE) loss:
\begin{equation}
    \mathcal{L}_{\text{est}} = \frac{1}{T} \sum_{t=1}^{T} \left\| \mathbf{v}_t^{\text{est}} - \mathbf{v}_t \right\|_2^2
\end{equation}

For safe obstacle avoidance in cluttered environments, we formulate an obstacle avoidance loss $\mathcal{L}_{\text{obs}}$ comprising two complementary components: a clearance loss based on a quadratic barrier, and a severe collision penalty loss using a softplus function, defined as

\begin{equation}
    \mathcal{L}_{\text{obs}} = \frac{1}{T} \sum_{t=1}^{T} v_{\text{rel}} \max(1 - (d_t-r),0)^2 + \beta_1 \ln(1+e^{-\beta_2 (d_t-r)})
\end{equation}
where $d_t$ denotes the distance from the center of the agent to the nearest obstacle point, $r$ is the radius of the quadrotor, $\beta_1$ weights the softplus collision penalty, $\beta_2$ controls the sharpness of the penalty near the obstacle boundary, and $v_{\text{rel}}$ scales the loss according to the relative obstacle approach rate.

Finally, $\mathcal{L}_\text{acc}$ and $\mathcal{L}_\text{jerk}$ regularize control smoothness by penalizing acceleration magnitude and its temporal variation (jerk), respectively, discouraging aggressive or oscillatory control commands:
\begin{equation}
\begin{aligned}
        \mathcal{L}_{\text{acc}} &= \frac{1}{T} \sum_{t=1}^{T} \|\mathbf{a}_t\|_2^2 \\
        \mathcal{L}_{\text{jerk}} & = \frac{1}{T-1} \sum_{t=1}^{T-1} \left\| \frac{\mathbf{a}_{t+1} - \mathbf{a}_t}{\Delta t} \right\|_2^2
\end{aligned}
\end{equation}

\section{Experiments}

\subsection{Event Data Preprocessing}
\label{sec:preprocessing}

To make real-world event data compatible with our simulator's representation, raw asynchronous events from the event camera are converted into the same frame-based, low-resolution format produced by our simulator. This preprocessing consists of a temporal aggregation step followed by a spatial downsampling step.

Raw events $e_k = (t_k, x_k, y_k, p_k)$ streamed from the camera are accumulated over a fixed time window $\Delta t = 1/15\,\text{s}$, matching the policy's control rate. For each raw pixel $(x,y)$, we aggregate the polarities of all events falling within the window by majority vote:
\begin{equation}
    \text{Event}_{x,y} = \text{sgn}
         \big(\sum_{e\in E}\mathbf{1}(e_p=+1) - \sum_{e\in E}\mathbf{1}(e_p=-1) \big)
\end{equation}
where $E$ denotes the event stream within the accumulation window. This yields a full-resolution ($640\times480$) event frame.

The full-resolution event frame is downsampled by a factor of 10 to $64\times48$, matching our simulator's rendering resolution, by aggregating pixel values within each $10\times10$ block and thresholding the result. This two-stage aggregation additionally suppresses sensor noise and task-irrelevant details in real event streams, reducing the sim-to-real gap.

\subsection{Offline Evaluation}

\begin{figure*}[t]
    \centering
    \begin{minipage}[b]{0.54\textwidth}
        \centering
        \begin{subfigure}[b]{0.48\linewidth}
            \centering
            \includegraphics[width=\linewidth]{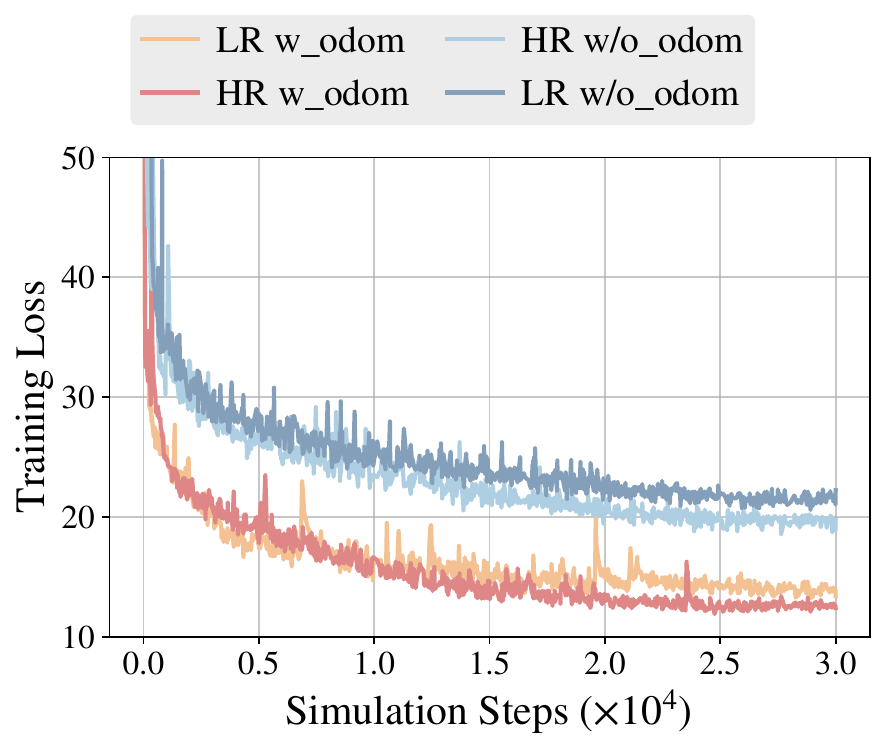}
            \caption{Training loss}
            \label{fig:loss}
        \end{subfigure}
        \hspace{0\textwidth}
        \begin{subfigure}[b]{0.48\linewidth}
            \centering
            \includegraphics[width=\linewidth]{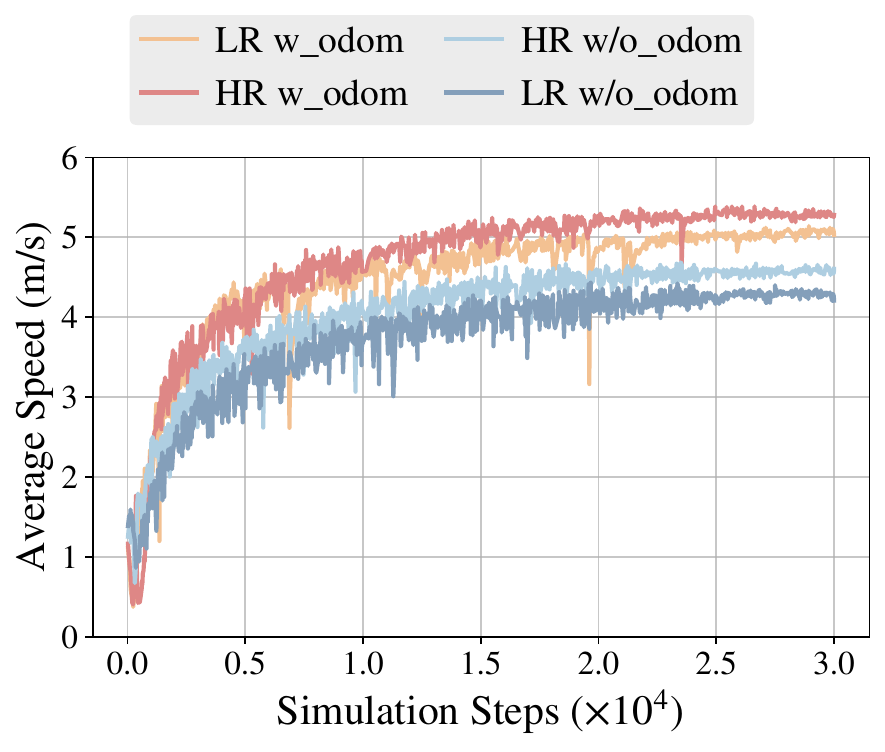}
            \caption{Average flight speed}
            \label{fig:speed}
        \end{subfigure}
        \captionof{figure}{Training loss and average flight speed over the course of training, for the odometry-assisted and odometry-free settings.}
        \label{fig:training_curves}
    \end{minipage}
    \hspace{0.01\textwidth}
    \begin{minipage}[b]{0.398\textwidth}
        \centering
        \includegraphics[width=\linewidth]{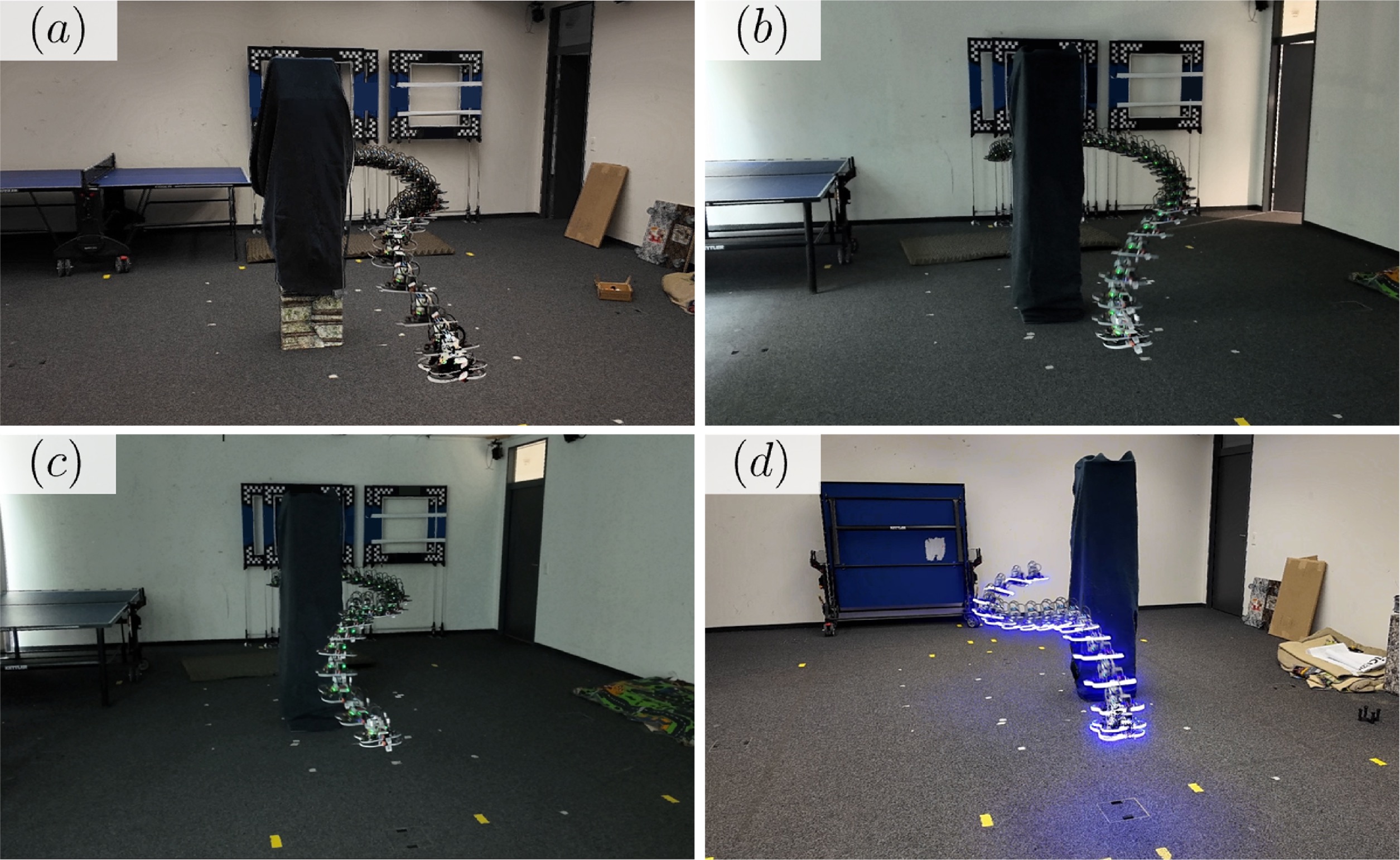}
        \captionof{figure}{Real-world flight trajectories under varying illumination and odometry conditions.}
        \label{fig:trajs}
    \end{minipage}
\end{figure*}
\begin{figure*}[t]
    \centering
    \begin{subfigure}[b]{0.2973\linewidth}
        \centering
        \includegraphics[height=4cm,width=\linewidth,keepaspectratio]{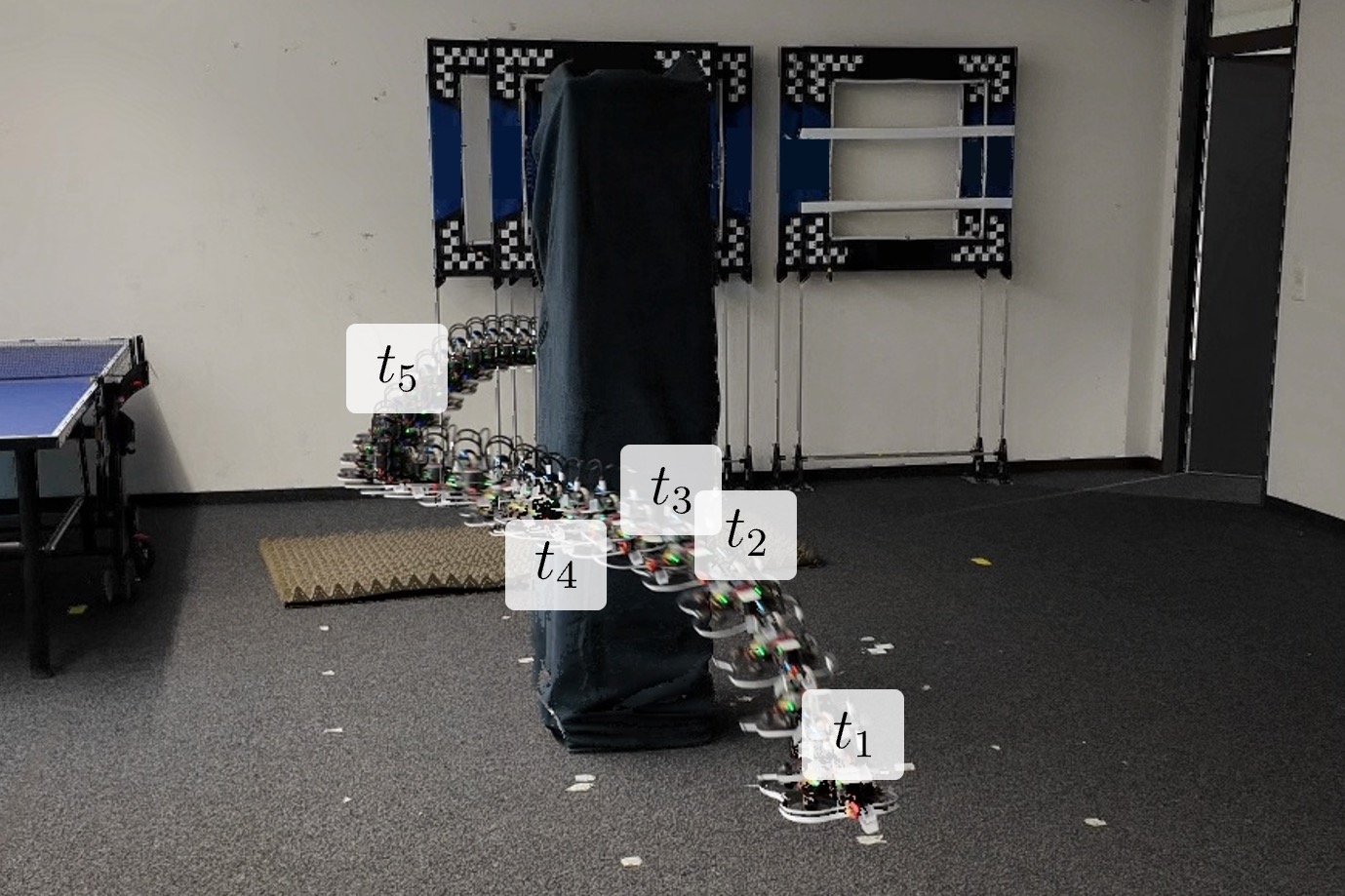}
        \caption{Scene overview}
        \label{fig:wide}
    \end{subfigure}%
    \hspace{1.5pt}%
    \begin{subfigure}[b]{0.13\linewidth}
        \centering
        \includegraphics[height=4cm,width=\linewidth,keepaspectratio]{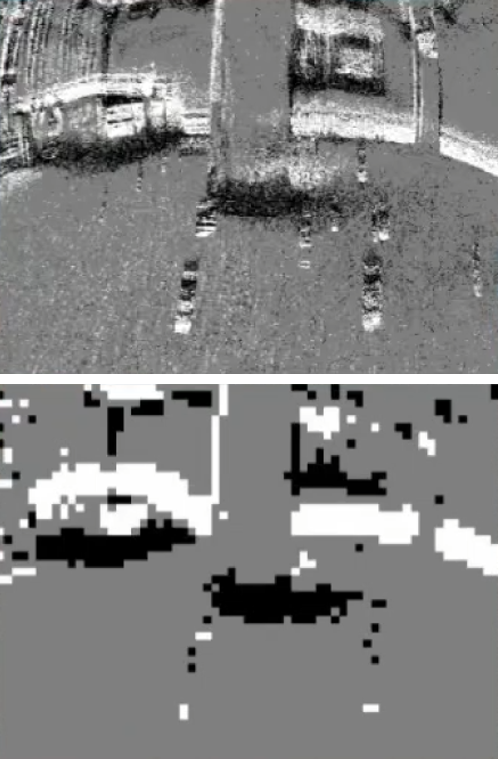}
        \caption{$t_1 = 1.8\,\text{s}$}
        \label{fig:t1}
    \end{subfigure}%
    \hspace{1.5pt}%
    \begin{subfigure}[b]{0.13\linewidth}
        \centering
        \includegraphics[height=4cm,width=\linewidth,keepaspectratio]{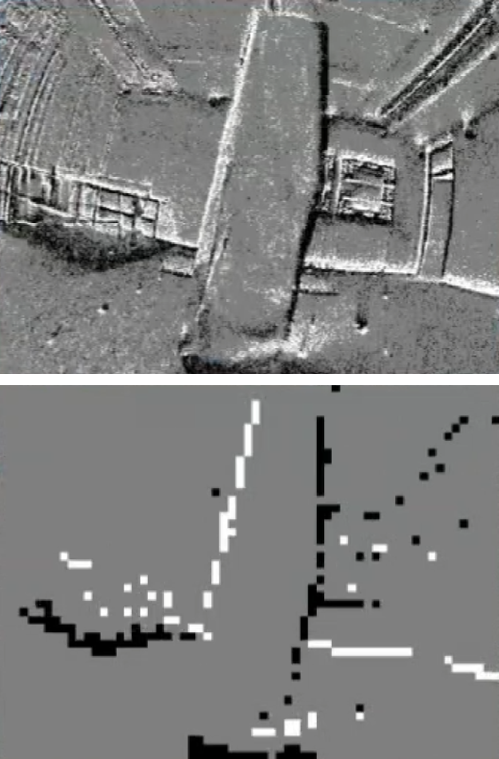}
        \caption{$t_2 = 2.1\,\text{s}$}
        \label{fig:t2}
    \end{subfigure}%
    \hspace{1.5pt}%
    \begin{subfigure}[b]{0.13\linewidth}
        \centering
        \includegraphics[height=4cm,width=\linewidth,keepaspectratio]{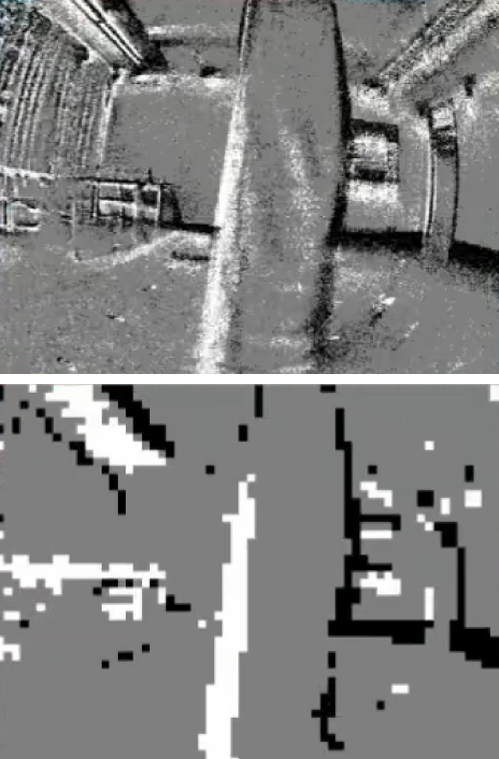}
        \caption{$t_3 = 2.3\,\text{s}$}
        \label{fig:t5}
    \end{subfigure}%
    \hspace{1.5pt}%
    \begin{subfigure}[b]{0.13\linewidth}
        \centering
        \includegraphics[height=4cm,width=\linewidth,keepaspectratio]{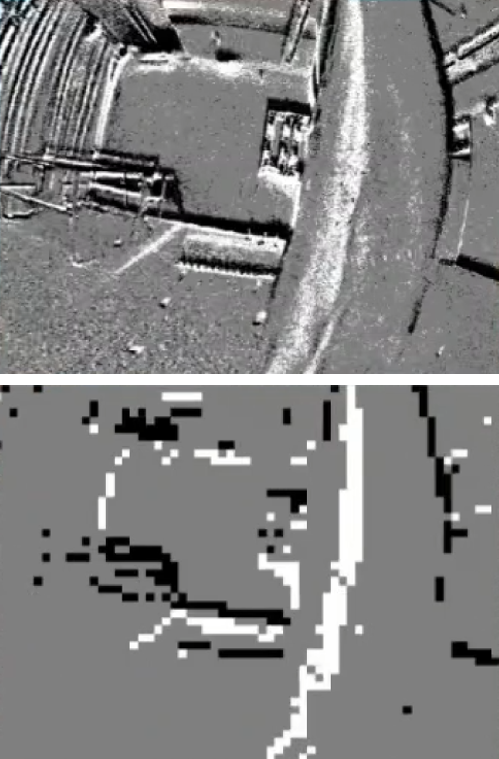}
        \caption{$t_4 = 3.4\,\text{s}$}
        \label{fig:t3}
    \end{subfigure}%
    \hspace{1.5pt}%
    \begin{subfigure}[b]{0.13\linewidth}
        \centering
        \includegraphics[height=4cm,width=\linewidth,keepaspectratio]{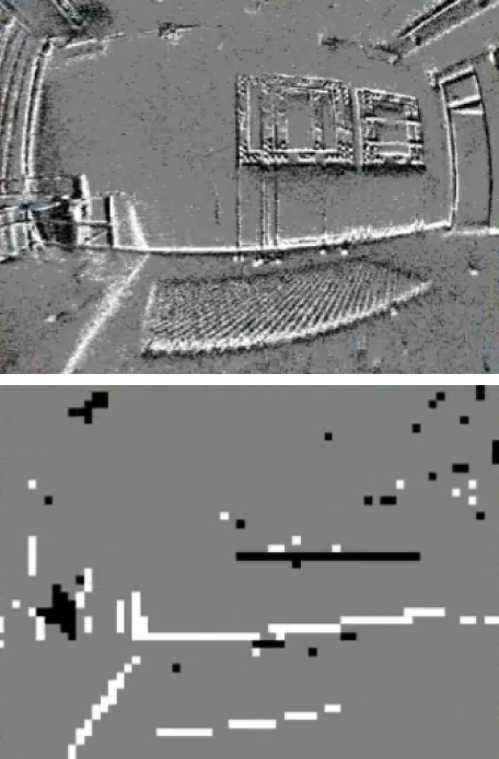}
        \caption{$t_5 = 4.7\,\text{s}$}
        \label{fig:t4}
    \end{subfigure}
    \caption{Flight trajectory across consecutive event camera timestamps. (a) shows the full scene, with the drone's position at each timestamp marked. For each timestamp $t_1$--$t_5$ in (b)--(f), the top row shows the raw event frames and the bottom row shows the corresponding downsampled event frames.}
    \label{fig:trajectory}
    \vspace{-0.4cm}
\end{figure*}
To assess the sim-to-real gap of our event simulator, we perform an offline evaluation through a supervised velocity estimation task independently of the control pipeline. The estimator adopts an architecture similar to our control policy, consisting of a CNN-based encoder and GRU, taking a sequence of event frames and the quadrotor's attitude as input and predicting velocity as output. We collected several trajectory datasets from our simulated environment and trained the estimator entirely on this simulated event data, without access to any real-world event data.

For real-world evaluation, we selected two representative sequences from the publicly available UZH-FPV Drone Racing Dataset~\cite{Delmerico19icra}. Fig.~\ref{fig:offline_eval} compares the predicted and ground-truth velocities. Despite being trained exclusively on simulated events, the predictions on real-world data closely tracked the ground-truth velocity trends, indicating that our simulator's event representation can transfer to real event data without any fine-tuning.
\subsection{Additional Details}
\label{sec:impl_details}

We optimize the policy with AdamW (learning rate $1\times10^{-3}$,
cosine-annealed to $1\times10^{-5}$, weight decay $0.01$) with
temporal gradient decay $\alpha=0.4$. The moving-average velocity
$\bar{\mathbf{v}}_t$ is computed over a $2$\,s window, and the
obstacle loss uses $\beta_1=2.5$ and $\beta_2=32$. The loss weights are $\lambda_\text{vel}=1.0$, $\lambda_\text{est}=100$, $\lambda_\text{obs}=3.0$, $\lambda_\text{acc}=0.01$, and $\lambda_\text{jerk}=0.001$. For the policy, the event frame is average-pooled to $16\times12$ and processed by
three convolutional layers with $32$, $64$, and $128$ filters
(kernel sizes $2$, $3$, and $3$; the first layer has stride $2$,
LeakyReLU activations), flattened, and linearly projected to a
$192$-dimensional feature. The target velocity and the state are concatenated and linearly projected to $192$ dimensions, and added to the image feature, and the combined feature is fed into a GRU cell whose
hidden state predicts the acceleration command and the velocity
estimate through a fully connected layer ($6$ outputs).
\subsection{Training Result}
\label{sec:training}

We trained our control policy on a laptop equipped with an NVIDIA RTX 5060 GPU, rendering event frames at a resolution of $64\times48$ and collecting 128 trajectories of 150 timesteps per iteration at a control interval of $1/15$~s.
Under this setup, the policy converged in approximately 3 hours, demonstrating the effectiveness of our differentiable simulation pipeline for policy training without reliance on large-scale compute infrastructure.

Fig.~\ref{fig:training_curves} shows the training loss and average flight speed over the course of training, across both odometry settings and simulator resolutions.
At both resolutions, the odometry-assisted policy converges faster and reaches a higher final average speed than its odometry-free counterpart.
This is expected, as the odometry-free policy must implicitly infer velocity-relevant cues from event frames alone, without the direct feedback available in the odometry-assisted setting.

We additionally compared our selected low resolution ($64\times48$) against a higher resolution ($96\times72$), which took approximately twice as long to train.
Despite this added cost, the higher-resolution policy achieved only marginal improvements in training loss and average speed over the low-resolution one, indicating that our selected resolution already provides sufficient information for effective policy learning.

\subsection{Sim-to-Real Transfer}
\label{sec:sim2real}
\subsubsection{Experiment Setup}
We evaluated our learned policy through zero-shot deployment on a real-world quadrotor platform. The quadrotor is equipped with a DVXplorer Micro event camera, which provides VGA ($640\times480$) resolution event output with up to 110~dB dynamic range and sub-millisecond latency.
This high dynamic range enables reliable perception under illumination conditions where conventional frame-based cameras typically fail. Real-time policy inference is performed onboard using an NVIDIA Jetson Orin NX at a fixed control rate of 15~Hz. Our policy predicts acceleration commands, which are converted into collective thrust and angular rate setpoints and tracked by a low-level attitude controller onboard the quadrotor. We model the flight controller response as first-order systems in our simulation, with delay coefficients identified via system identification on the real platform. In particular, the desired heading is derived from the direction toward the goal and tracked through the first-order system response.

\subsubsection{Real-World Flights}
Real-world experiments were conducted in a single indoor environment containing one static obstacle. The quadrotor was commanded to fly toward a goal location on the opposite side of the obstacle at a target velocity using onboard event perception. We evaluated our policy under varying illumination and odometry conditions, as well as randomized obstacle positions. Despite these variations across trials, the policy consistently generates smooth avoidance trajectories that adapt to the change.

For the odometry-assisted setting, results are shown in Fig.~\ref{fig:trajs}(a)--(c), where the policy has access to both attitude and velocity as state input. With odometry removed, given in Fig.~\ref{fig:image1} (top) and Fig.~\ref{fig:trajs}(d), the policy instead relies solely on onboard attitude information without any velocity input, yet still produces stable avoidance behavior comparable to the odometry-assisted setting. We evaluated the real-world flight outcomes under different odometry conditions, as summarized in Table~\ref{tab:odom_success}.

We further tested an extremely dark condition (Fig.~\ref{fig:image1}, bottom), with illuminance as low as 5.8\,lux. Performance remained unaffected due to the high dynamic range of the event camera, in contrast to conventional frame-based cameras, which typically fail under such conditions.

To examine the onboard perception underlying these maneuvers, Fig.~\ref{fig:trajectory} visualizes representative event views at several time steps along the flight trajectory. Fig.~\ref{fig:trajectory} (a) shows the external view with these time steps annotated, while (b)--(f) show the corresponding raw (top row) and downsampled (bottom row) event frames. Despite the substantially reduced resolution, obstacle edges remain clearly discernible in the downsampled event frames, confirming that our low-resolution rendering design preserves the information necessary for reliable obstacle avoidance.

\subsection{Baseline Comparison}
\begin{table}[b]
\vspace{-0.3cm}
    \centering
    \caption{Real-world flight outcomes under different odometry conditions.}
    \label{tab:odom_success}
    \begin{tabular}{lcc}
        \toprule
        Condition & Successful & Total Trials \\
        \midrule
        Odometry-assisted & 7 & 8 \\
        Odometry-free      & 3 & 3 \\
        \bottomrule
    \end{tabular}
    \vspace{0.3cm}
\end{table}
\begin{figure}[t]
    \centering
    \begin{subfigure}[b]{0.6\linewidth}
        \centering
        \includegraphics[width=\linewidth]{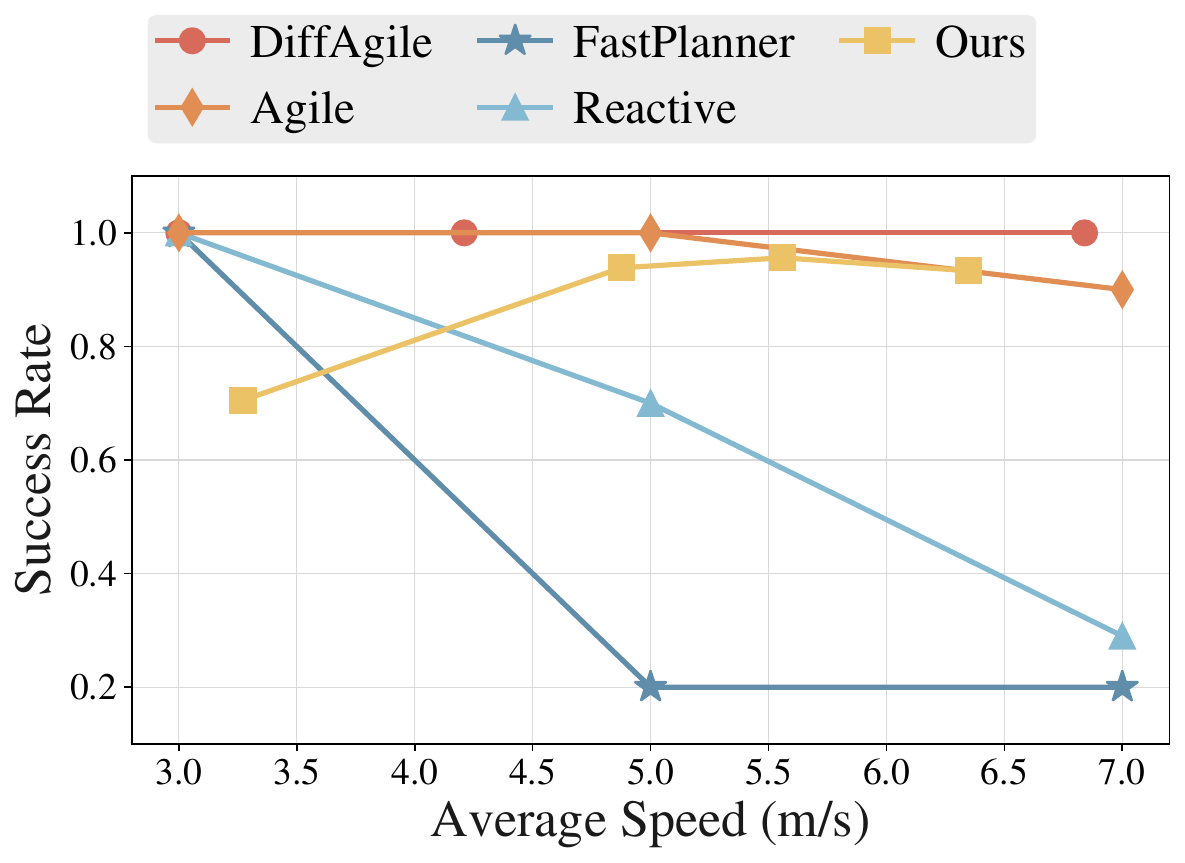}
        \caption{Success rate vs.\ average speed}
        \label{fig:baseline_curve}
    \end{subfigure}
    \hfill
    \begin{subfigure}[b]{0.37\linewidth}
        \centering
        \includegraphics[width=\linewidth]{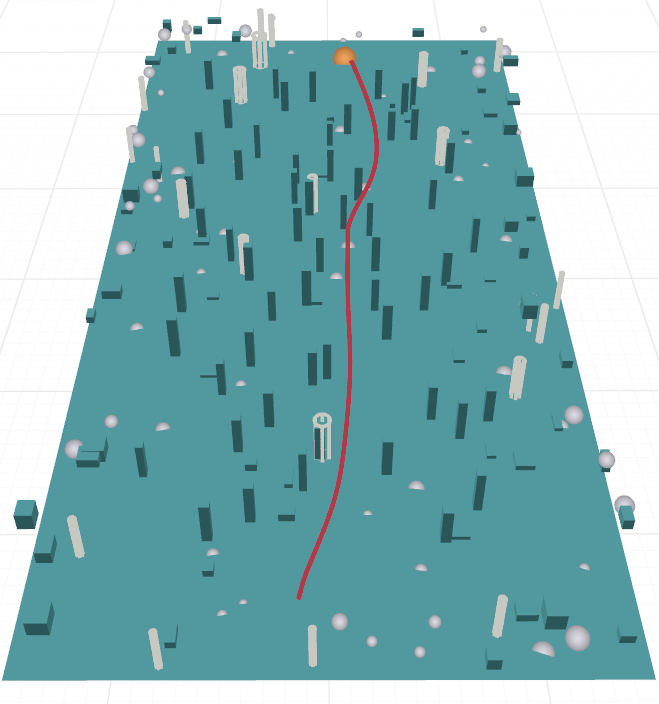}
        \caption{Test environment}
        \label{fig:baseline_env}
    \end{subfigure}
    \caption{Representative comparison between our event-based policy and the depth-based baselines reported in~\cite{zhang2025learning}. (a) Success rate across average flight speed. (b) Top-down view of the evaluation environment and the drone's trajectory.}
    \label{fig:baseline}
    \vspace{-0.8cm}
\end{figure}

For baseline comparison, we reference the results reported for the depth-based method DiffAgile~\cite{zhang2025learning}, which has been shown to achieve state-of-the-art performance compared to baselines such as FastPlanner~\cite{zhou2019robust}, Agile~\cite{antonio2021agile}, and Reactive~\cite{florence2020integrated}.
The result is given in Fig.~\ref{fig:baseline}. Under this framing, our event-based policy trailed DiffAgile at low speed but briefly approached it at higher tested speeds, despite relying \emph{solely on monocular event input without depth}.
This result aligns with the underlying event generation mechanism, as slower motion triggers fewer events, providing less perceptual information to the policy at low speed.
Beyond flight performance, DiffAgile~\cite{zhang2025learning} demonstrated improved sample efficiency over PPO, motivating our use of differentiable policy optimization.
In our event-based setting, the policy converges in approximately 3 hours on an NVIDIA RTX 5060 laptop GPU, demonstrating practical training efficiency.
While this supports the effectiveness of our training framework, we do not directly compare its sample efficiency against PPO.

It is also worth noting that depth-based methods rely on depth sensors that, in the real world, suffer from degradation such as motion blur, which typically worsens under low-light conditions due to longer exposure times.
Event cameras are less sensitive to motion blur, suggesting our approach may hold greater advantages in real-world low-light deployment that are not reflected in this simulated comparison.

\subsection{Simulator Efficiency Comparison}
\label{sec:sim_efficiency}
\begin{figure}
    \centering
    \includegraphics[width=1\linewidth]{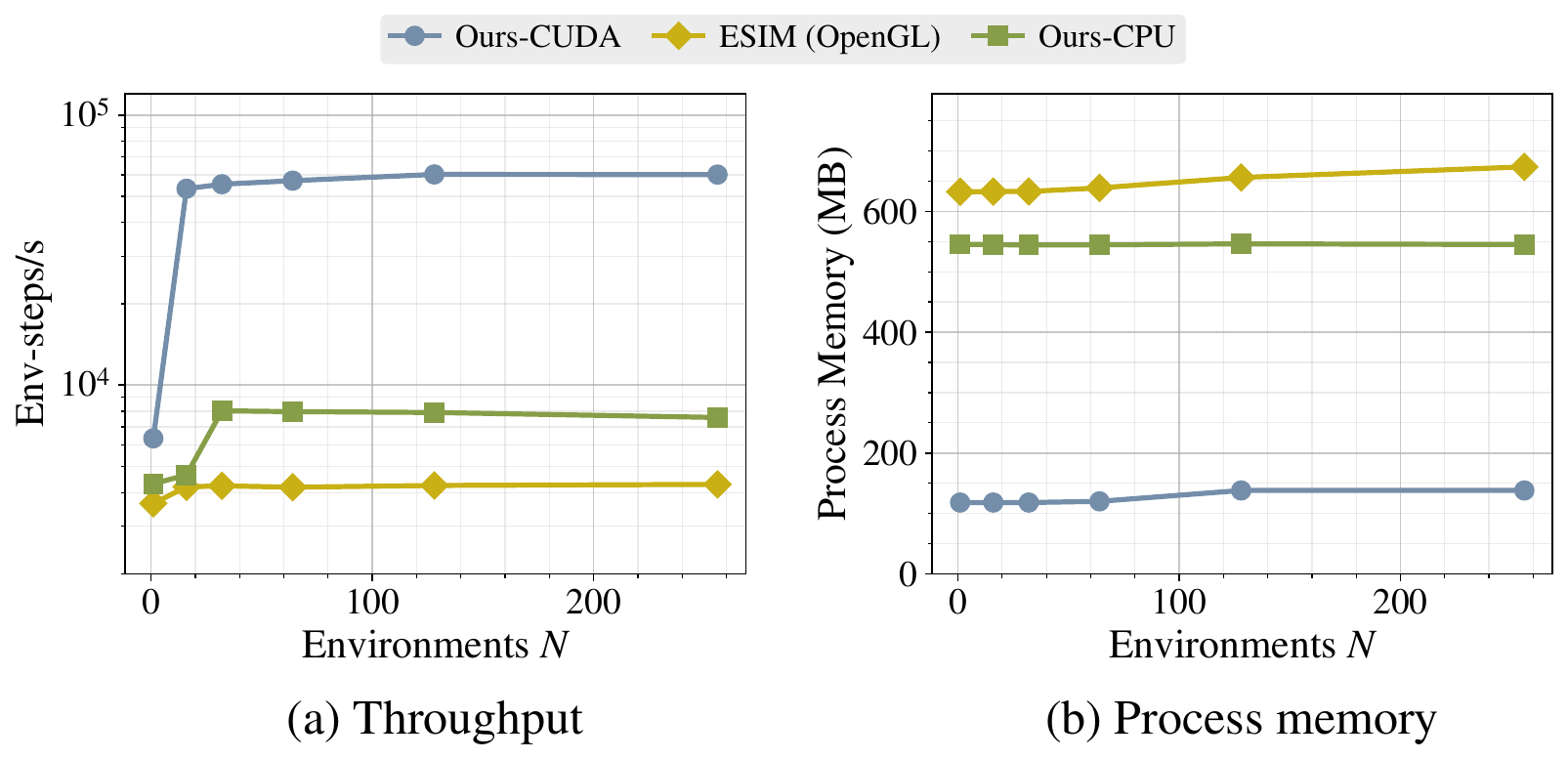}
    \caption{Simulation throughput and memory vs. number of parallel environments $N$. (a) Throughput in environment steps per second. Our method's advantage over ESIM widens with increasing $N$, until the GPU's streaming multiprocessors are saturated. (b) Process memory footprint. Our method's GPU memory grows only marginally with $N$, whereas ESIM's host memory grows with $N$ due to its per-environment simulator state.}
    \label{fig:scaling}
    \vspace{-0.6cm}
\end{figure}

A central motivation for our CUDA-based event simulator is to remove the rendering bottleneck that limits large-scale policy training. This section evaluates our simulator against other event simulators on computational efficiency.

We benchmarked our simulator against ESIM~\cite{rebecq2018esim}, which couples an OpenGL renderer with a CPU-based event generation core, and against a multi-threaded CPU-only implementation of our own simulator. All three methods render the same $64\times48$ pinhole camera view of an identical scene
containing 200 randomly placed obstacles, following the same
camera trajectory sampled at 15\,Hz. The number of parallel environments $N$ ranges from 1 to 256.
All measurements were collected on the same NVIDIA RTX 5060 laptop GPU used for policy training.
Fig.~\ref{fig:scaling} summarizes the results. Our simulator's throughput rose sharply with $N$ and reached 60{,}248 environment steps per second at $N=128$ once the GPU's compute units were saturated. Compared to 4{,}249 steps per second for ESIM, our simulator achieved a speedup of approximately $14\times$. Furthermore, at $N=128$, our simulator used only 138\,MB of GPU memory, less than the host memory footprint of ESIM (656\,MB). These results indicate that our simulator's efficiency, rather than the training procedure alone, is what enables efficient policy convergence.

\section{Conclusion}

\label{sec:discussion}

In this work, we present an end-to-end learning framework that directly maps event frames to control commands, and validate it through zero-shot deployment on a real-world quadrotor platform in both odometry-assisted and odometry-free settings.
Our results confirm that a CUDA-based event simulator, combined with a differentiable simulation pipeline, enables efficient policy training, and that the resulting policy generalizes reliably to real-world flight, including under low-light conditions.
Despite these results, several limitations remain.
Our simulator does not model surface texture on obstacles, which limits the visual realism of the rendered grayscale and event frames and may contribute to a sim-to-real gap.
Furthermore, our current framework considers only static obstacles. Extending the simulation to incorporate dynamic obstacles during training could be the next step toward more general autonomous flight in unstructured, real-world environments.

\bibliographystyle{ieeetr}
\bibliography{references}

\end{document}